\documentclass[lettersize,journal]{IEEEtran}
\usepackage{amsmath,amsfonts}
\usepackage{algorithmic}
\usepackage{algorithm}
\usepackage{array}
\usepackage[caption=false,font=normalsize,labelfont=sf,textfont=sf]{subfig}
\usepackage{textcomp}
\usepackage{stfloats}
\usepackage{url}
\usepackage{verbatim}
\usepackage{graphicx}
\usepackage{cite}
\usepackage{array}
\usepackage{multirow}
\usepackage{makecell}
\usepackage{booktabs}
\usepackage{flushend}

\newcolumntype{C}[1]{>{\centering\arraybackslash}m{#1}}
\newcolumntype{L}[1]{>{\raggedright\arraybackslash}m{#1}}

\begin{document}

\title{
Task-Oriented Sensing and Covert Transmissions for Collaborative Multi-AUV Systems}

\author{Xueyao~Zhang, Bo~Yang, Chenyang Yan, Xuelin Cao, Zhiwen Yu, Bin Guo,\\  Samson Lasaulce, George C. Alexandropoulos, and M\'erouane Debbah

\thanks{


X. Zhang, B. Yang, C. Yan, and B. Guo are with the School of Computer Science, Northwestern Polytechnical University, Xi'an, Shaanxi, 710129, China (email:  yang$\_$bo, guob@nwpu.edu.cn). 

X. Cao is with the School of Cyber Engineering, Xidian University, Xi'an, Shaanxi, 710071, China (email: caoxuelin@xidian.edu.cn). 

Z. Yu is with the School of Computer Science, Northwestern Polytechnical University, Xi'an, Shaanxi, 710129, China, and Harbin Engineering University, Harbin, Heilongjiang, 150001, China (email: zhiwenyu@nwpu.edu.cn).

S. Lasaulce is with Khalifa University, UAE, and CRAN, Nancy, France (email: samson.lasaulce@univ-lorraine.fr).

G. C. Alexandropoulos is with the Department of Informatics and Telecommunications, National and Kapodistrian University of Athens, 16122 Athens, Greece (email: alexandg@di.uoa.gr). 

M. Debbah is with the Research Institute for Digital Future, Khalifa University, 127788 Abu Dhabi, UAE (email: merouane.debbah@ku.ac.ae) and also with CentraleSupelec, University Paris-Saclay, 91192 Gif-sur-Yvette, France.


}
}

\markboth{Journal of \LaTeX\ Class Files,~Vol.~14, No.~8, August~2021}%
{Shell \MakeLowercase{\textit{et al.}}: A Sample Article Using IEEEtran.cls for IEEE Journals}


\maketitle

\begin{abstract}
In underwater covert cooperative missions, autonomous underwater vehicles (AUVs) often cannot rely on active sonar to continuously obtain complete information, since active sensing and frequent communications increase exposure risk. As a result, AUVs primarily rely on passive observation, which yields incomplete local perception and limited task efficiency. Although underwater acoustic communications can mitigate this limitation through information sharing, they are simultaneously constrained by long delays, severe interference, low reliability, and the risk of covert exposure. Existing communications-oriented multi-agent reinforcement learning (MARL) studies often model communication as an ideal information flow, whereas traditional communication optimization primarily focuses on link-level performance. However, both are insufficient to characterize the actual contribution of perceptual information to cooperative tasks under realistic covert physical communication conditions. This paper proposes a Sensed-information Value Realization Multi-Agent Reinforcement Learning (SVR-MARL) framework that leverages practical information to characterize information utility for cooperative tasks and learns distributed cooperative policies under realistic communication and covert constraints. Through a case study of covert multi-AUV cooperative localization and tracking, the potential of the proposed framework to improve collaborative task efficiency while reducing unnecessary communication and exposure risks is demonstrated.

\end{abstract}


\section{Introduction}
\IEEEPARstart{A}{s} sixth-generation (6G) communications evolve from conventional data transmission toward an integrated network paradigm that deeply combines communication, sensing, computing, and intelligence, the application boundaries of network systems are continuously expanding from traditional terrestrial spaces to more open and complex ocean environments \cite{underwater}. Autonomous underwater vehicles (AUVs), as key platforms for marine environmental perception and autonomous operations, are widely used for tasks such as ocean monitoring, underwater security, resource exploration, and target tracking \cite{AUV}. However, in complex underwater environments, a single AUV often cannot accomplish large-scale, long-duration, and highly dynamic tasks based solely on limited observations. Consequently, multi-AUV cooperation has gradually become an important approach for improving environmental perception capability and mission efficiency. Compared with terrestrial wireless environments, underwater spaces inherently suffer from severe propagation attenuation, long propagation delays, narrow bandwidth, and strong dynamics, which pose significant challenges to information exchange and cooperative decision-making.

In complex underwater environments, multi-AUV systems usually rely on active sensing or information exchange to acquire richer environmental and target information. Underwater perception commonly relies on sonar systems. Among them, active sonar can obtain range and other target-related information by actively transmitting acoustic signals. However, its high-power acoustic radiation can be directly detected by distant devices, making it difficult to satisfy the survivability requirements of covert missions. In contrast, passive sonar offers greater concealment and is therefore more suitable as the primary sensing approach for covert underwater tasks. However, passive observations usually provide only limited local bearing information, making it difficult for an individual AUV to accurately recover the target state and potentially leading to accumulated perception uncertainty and unstable localization. To mitigate this ambiguity, multiple AUVs can exploit spatially complementary information from different locations and exchange local observations and cooperative states via limited underwater acoustic communication, thereby improving overall environmental perception and cooperative capabilities. Nevertheless, inappropriate communications may also introduce exposure risks. Therefore, a fundamental challenge is supporting cooperative information exchange under low-detectability constraints.

Multi-Agent Deep Reinforcement Learning (MADRL) provides an end-to-end policy-learning paradigm for cooperative tasks, in which agents learn collaborative behaviors through environmental interactions under shared objectives. Because each agent can make decisions only from local observations during distributed execution, MADRL incorporates communication mechanisms to ease coordination challenges arising from partial observability. Representative studies such as IC3Net \cite{IC3} have demonstrated that inter-agent information exchange can improve cooperative decision-making. Specifically, these methods typically model communication as an information-exchange channel within the policy network, allowing agents to learn when to communicate and how to use the information they receive. However, the ``communication'' considered in these works is often highly idealized. Messages are modeled as abstract information flows, and practical constraints on transmission and decoding in real-world physical systems are rarely considered.

\begin{figure}[t]
	\centering
	\includegraphics[width=1.01\columnwidth]{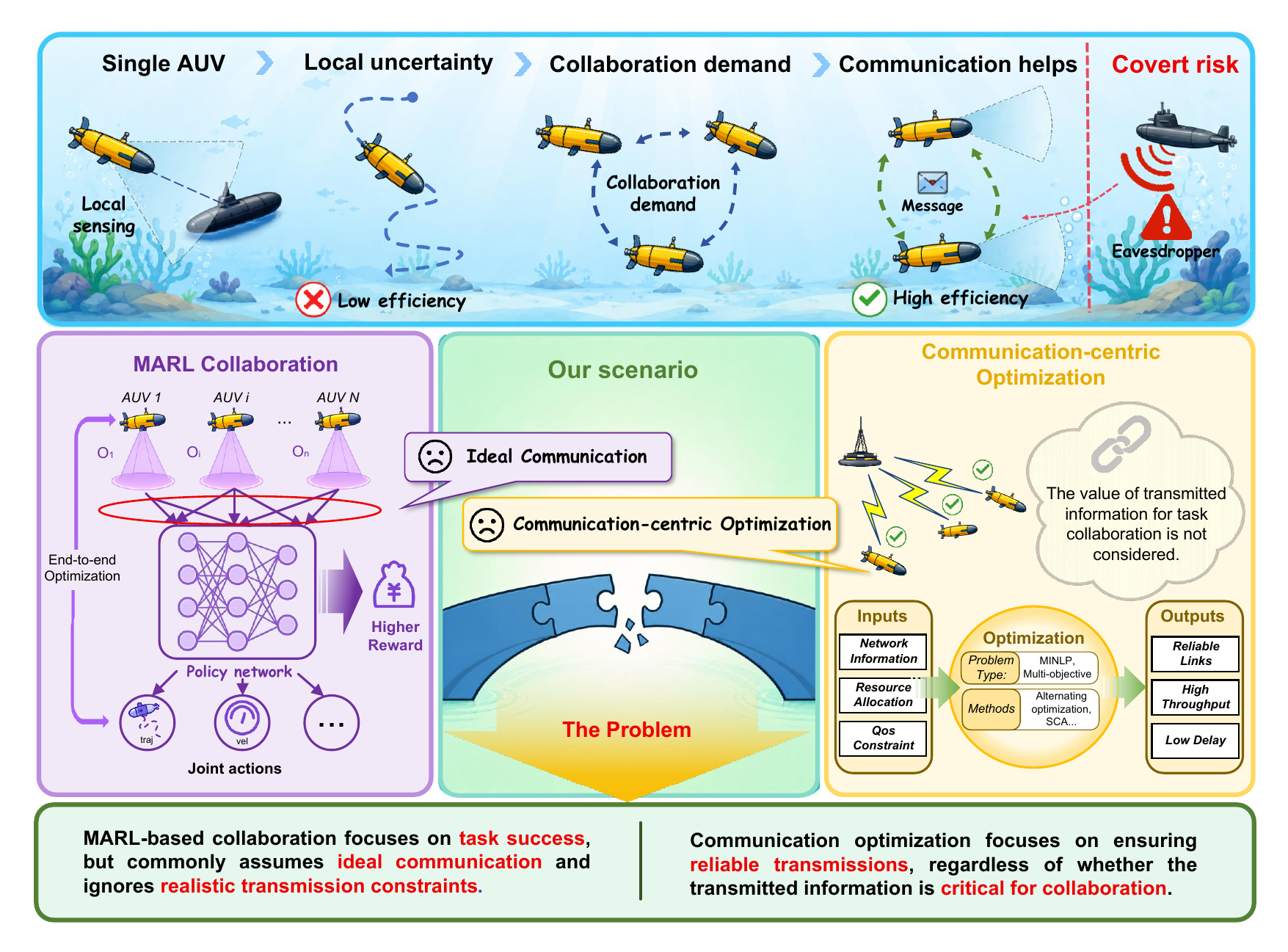}
	\caption{Task motivation and limitations of MARL-based collaboration and communication-centric optimization for covert multi-AUV collaboration.}
    \label{p1}
\end{figure}

Research on underwater wireless communications has primarily focused on ensuring information transmission performance over constrained acoustic links. Existing studies commonly formulate communication as a link-quality or resource-allocation problem and improve transmission reliability and communication efficiency through optimization or learning-based approaches. Covert communication studies \cite{covert01} further incorporate low detectability into the design. However, their primary objective remains improving data transmission rates under covert constraints. Overall, current research largely focuses on ensuring reliable communication, while paying limited attention to whether the transmitted information contributes to cooperative tasks in such harsh environments. In our covert collaboration scenario, this may lead to redundant communication, inefficient resource use, and unnecessary exposure risks.

The aforementioned studies characterize “information interaction” in cooperative communication and ``link reliability" in physical communication, respectively. However, they remain insufficient for the more practical underwater multi-agent covert collaboration scenario considered in this work. First, more realistic communication transmission processes need to be considered. On the other hand, in harsh, resource-constrained environments, communication with higher collaborative value should be prioritized. Motivated by these challenges, this work focuses on the following three questions:
\begin{itemize}
    \item \textbf{Q1}: In covert collaborative scenarios, how can we determine whether a piece of local perceptual information truly contributes to the team task?
    \item \textbf{Q2}: When messages must be transmitted through real underwater acoustic links, how can we characterize the effects of propagation, decoding, and covert requirements on their collaborative value?
    \item \textbf{Q3}: Under distributed partial-observation settings, how can AUVs learn when to communicate and which information is worth transmitting to avoid communication interference and unnecessary exposure?
\end{itemize}

\section{Overview of Multi-agent Underwater Collaboration}
\subsection{Information-Flow Communication in MARL}
Communication-enabled MARL has been widely studied to address partial observations in distributed multi-agent systems. Under the centralized training and decentralized execution (CTDE) paradigm, agents can learn collaborative behaviors toward shared task objectives. Representative methods include value decomposition methods (e.g., VDN and QMIX) and policy optimization methods (e.g., MADDPG and MAPPO), both of which have demonstrated strong cooperative capabilities in tasks such as cooperative navigation and target tracking. However, during distributed execution, each agent typically relies only on local observations and cannot directly access other agents’ states, intentions, or global environmental information. This limited information becomes a key bottleneck for efficient collaboration.

To alleviate collaboration difficulties arising from partial observations, researchers have incorporated communication mechanisms into MADRL, in which agents exchange encoded messages to supplement local observations and improve joint decision-making. These messages usually represent local observations, task intentions, or compact state summaries. Early works such as CommNet \cite{CommNet} modeled inter-agent communication as a continuous, differentiable process and learned communication protocols and cooperative policies end-to-end under idealized communication assumptions. IC3Net \cite{IC3} introduced a gating mechanism for selective communication, enabling agents to learn when to transmit messages. TarMAC \cite{TarMac} further adopted attention-based targeted communication, allowing agents to decide what messages to send and whom to communicate with. These studies show that learned communication can mitigate incomplete local information and improve collaboration efficiency.

However, the ``communication" in the aforementioned methods essentially remains within the scope of information-flow communication. Their primary focus is on how messages are exchanged among policy networks, rather than on how information is physically transmitted between real-world entities. During training and execution, inter-agent messages are typically treated as directly accessible abstract information flows, implicitly assuming that messages can be delivered in a timely, reliable, and low-cost manner. As a result, practical factors such as propagation delay, link interference, decoding reliability, and covert communication requirements are rarely considered. When these methods are applied to covert underwater collaboration, such idealized assumptions create a clear gap between learned cooperation and physical realizability.

\subsection{Communication Optimization for Covert Networks}
Research on underwater wireless communications has long focused on reliable transmission over complex acoustic links. Since acoustic propagation is strongly affected by transmission distance, signal frequency, ocean conditions, and node mobility, existing studies typically optimize link reliability, data transmission rate, and energy efficiency \cite{COMST}. At the physical layer, techniques such as adaptive modulation and coding, channel estimation, and equalization are widely employed to adapt to time-varying channel conditions, mitigate inter-symbol interference, and reduce bit error rates \cite{estimation}. At the link layer, MAC scheduling and power control mitigate conflicts and resource competition over shared acoustic channels \cite{link}, thereby improving network throughput and energy utilization efficiency.

In covert communications, research has further considered information transmission under low-detectability requirements. Existing works \cite{covert02} commonly jointly optimize transmission power, energy consumption, AUV trajectory, and transmission strategies to improve communication reliability and performance while satisfying covert constraints. Nevertheless, their objectives still primarily serve communication itself, i.e., reliable and covert data transmission. For multi-AUV covert collaboration tasks, link availability does not imply that a transmission is worthwhile, since some messages may contribute little or even detract from team collaboration. Blind transmissions may instead waste resources, cause link conflicts, and miss valuable covert communication opportunities.

\begin{table*}[t]
\centering
\caption{Comparison of different communication approaches.}
\label{tab:comparison}
\renewcommand{\arraystretch}{1.35}
\setlength{\tabcolsep}{5pt}
\footnotesize

\begin{tabular}{
>{\centering\arraybackslash}m{2.6cm}
>{\centering\arraybackslash}m{4.0cm}
>{\centering\arraybackslash}m{3.0cm}
>{\centering\arraybackslash}m{3.0cm}
>{\centering\arraybackslash}m{3.0cm}
}
\hline
\textbf{Block} 
& \textbf{Representative Works} 
& \textbf{Realistic Channel Constraints} 
& \textbf{Covert Transmission} 
& \textbf{Task-oriented Value} \\
\hline

\multirow[c]{3}{*}{%
\parbox[c]{2.5cm}{\centering\textbf{MARL Collaboration\\with ideal\\communication}}}
& Sukhbaatar et al.~\cite{CommNet} 
& $\times$ & $\times$ & $\checkmark$ \\
& Singh et al.~\cite{IC3} 
& $\times$ & $\times$ & $\checkmark$ \\
& Das et al.~\cite{TarMac}, Niu et al.~\cite{MAGIC} 
& $\times$ & $\times$ & $\checkmark$ \\
\hline

\multirow[c]{4}{*}{%
\parbox[c]{2.5cm}{\centering\textbf{Communication-\\centric\\Optimization}}}
& Jiang~\cite{COMST} 
& $\checkmark$ & $\times$ & $\times$ \\
& Liang et al.~\cite{estimation} 
& $\checkmark$ & $\times$ & $\times$ \\
& Zhang et al.~\cite{wcnc} 
& $\checkmark$ & $\checkmark$ & $\times$ \\
& Chen et al.~\cite{covert02} 
& $\checkmark$ & $\checkmark$ & $\times$ \\
\hline

\textbf{Our work}
& SVR-MARL 
& $\checkmark$ & $\checkmark$ & $\checkmark$ \\
\hline
\end{tabular}
\end{table*}

\subsection{The Missing Gap}
Existing studies have improved multi-agent collaboration and physical communications from the perspectives of collaborative learning and transmission optimization, respectively. However, these approaches remain insufficient for covert underwater multi-AUV collaboration. On the one hand, MARL-based message passing typically lacks realistic underwater acoustic modeling, including propagation delays, interference, decoding reliability, and covert constraints. On the other hand, traditional communication optimization focuses on the physical transmission process and mainly aims to improve communication performance itself, without considering the task significance of the message content. Therefore, in covert collaboration scenarios, communication decisions should depend not only on link availability but also on whether the transmitted information can truly benefit team collaboration.

Motivated by this gap, this work focuses on learning AUV cooperative policies distributively under covert constraints to maximize overall collaborative efficiency. We introduce the concept of “value of sensed information” to characterize the potential contribution of local observations to collaborative tasks, as illustrated in Fig.~\ref{p1}. Compared with conventional communication schemes centered on link performance, we emphasize the actual impact of local perceptual information on team collaboration under realistic physical communication conditions, while jointly considering information value, link availability, and covert requirements in communication decision-making. From this perspective, covert communication is no longer a purely data-transmission problem but an information-selection process that serves collaborative perception and cooperative task execution.

\section{SVR-MARL Framework for Covert Collaboration}
\begin{figure}[t]
	\centering
	\includegraphics[width=1.0\columnwidth]{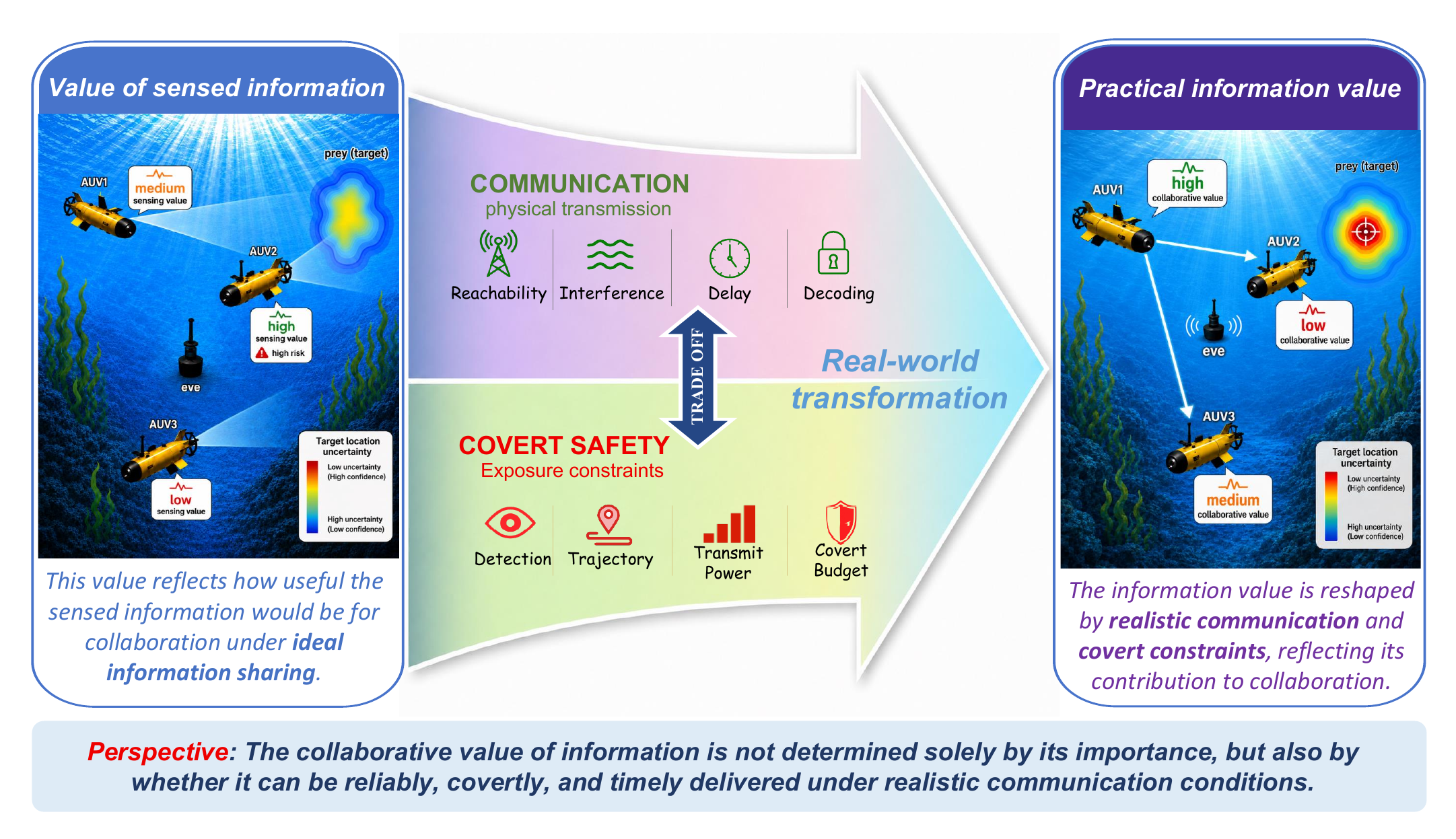}
	\caption{A perspective on realizing collaborative value under realistic communication and covert constraints.}
    \label{sensingv}
\end{figure}

In covert multi-AUV collaboration, communication decisions should consider not only the potential contribution of local sensed information to cooperative tasks, but also whether this contribution can be realized under realistic physical communication and covertness conditions. Motivated by this consideration, this section introduces the concepts of \textit{value of sensed information} and \textit{practical information value}. As illustrated in Fig.~\ref{sensingv}, the \textit{value of sensed information} describes the potential collaborative contribution of local observations before considering practical communication effects, such as providing complementary target observations, improving sensing geometry, or reducing team uncertainty. The \textit{practical information value} further reflects the collaborative utility that these potential contributions can realize after actual transmission, reliable reception, and task utilization. Consequently, link conditions, decoding outcomes, and exposure risks can alter the relative utility of different communication actions and reshape the probability distributions of transmission gating, transmit power, and other communication actions during policy learning. Accordingly, this section focuses on how task outcomes, together with realistic communication and covertness feedback, can optimize distributed communication and task policies, thereby shifting communication decisions from the potential collaborative contribution of sensed information toward the practical information value realizable under realistic conditions.

\subsection{Task Modeling and Action Decomposition}
In complex underwater environments, many tasks involve long durations, wide operational areas, high dynamics, and strong uncertainty. A single AUV often struggles to complete such missions independently because of limited sensing capability and restricted mobility. For example, in target tracking, cooperative navigation, area search, and pipeline inspection, multiple AUVs can achieve complementary perception through spatially distributed deployment, and further improve task efficiency through limited information exchange for state sharing, intention coordination, and cooperative execution.
Although different tasks may involve different objectives and execution actions, their decision-making processes can generally be divided into two basic aspects: how to exchange information that benefits collaboration, and how to perform task-oriented physical actions. Therefore, we extract their common structure and decompose each AUV’s policy into a communication phase and a task-execution phase, with the corresponding action expressed as $a_i={a_i^{\mathrm{comm}},a_i^{\mathrm{task}}}$.
Here, $a_i^{\mathrm{comm}}$ describes decisions related to information transmission, whose core role is to determine whether local sensing messages can be delivered over realistic underwater acoustic links reliably, on time, and covertly. Depending on the specific underwater communication model, $a_i^{\mathrm{comm}}$ may include transmission gating, transmit power, bandwidth, frequency, and access time slots. The task action $a_i^{\mathrm{task}}$ is oriented toward the specific cooperative objective and can be represented by speed, heading, depth, or task-interaction actions, depending on the task type.

\subsection{Coupling Mechanisms}
\subsubsection{Covert-Constrained Task Coupling}
For the covert communication problem considered here, we adopt probabilistic detection models for underwater covert communications to make covertness compatible with optimization and learning. Specifically, we model the eavesdropper's listening behavior as a binary hypothesis-testing problem and transform the low-detectability requirement into a tractable covertness condition using the KL divergence. Meanwhile, the receiver must also meet a decoding reliability requirement to ensure that teammates can reliably receive messages.
Based on this modeling, covertness couples the communication and task phases. On the one hand, transmission decisions and transmit power affect the received signal strength at the eavesdropper, thereby influencing exposure risk. On the other hand, the distance and relative geometry between an AUV and the eavesdropper affect the channel gain, meaning task actions can also indirectly influence covertness. Therefore, the covertness requirement serves as a key coupling factor between the communication and task phases. In essence, the communication action determines “at what cost information is transmitted,” while the task action determines “under what geometry the task and communication are performed.” These two phases should therefore be considered jointly under a unified objective of covert collaboration, rather than designed independently.

\subsubsection{Closed-Loop Coupling in POMG}
A Markov game can be viewed as the multi-agent extension of a Markov decision process. For underwater multi-AUV collaborative tasks, each AUV can only obtain local observations and cannot directly access the global state. Therefore, we formulate the problem as a partially observable Markov game (POMG). In this model, each AUV first forms its communication observation $o_i^{\mathrm{comm}}(t)$ based on its local observation, historical information, and received messages, and then generates the communication action $a_i^{\mathrm{comm}}(t)$. After passing through realistic underwater acoustic links, the communication may result in different outcomes, such as delayed arrival, decoding failure, or successful reception. These outcomes further change the information available to the AUV and affect the task-stage observation $o_i^{\mathrm{task}}(t)$. Subsequently, the AUV executes the task action $a_i^{\mathrm{task}}(t)$ based on the updated task observation, thereby driving the environment state transition.
This transition changes the spatial geometry, channel conditions, interference relationships, and covertness-related factors at the next time step, thereby further affecting the subsequent communication observation $o_i^{\mathrm{comm}}(t+1)$. Therefore, the communication and task phases form a closed-loop coupling in the POMG.
\begin{figure}[t]
	\centering
	\includegraphics[width=1.01\columnwidth]{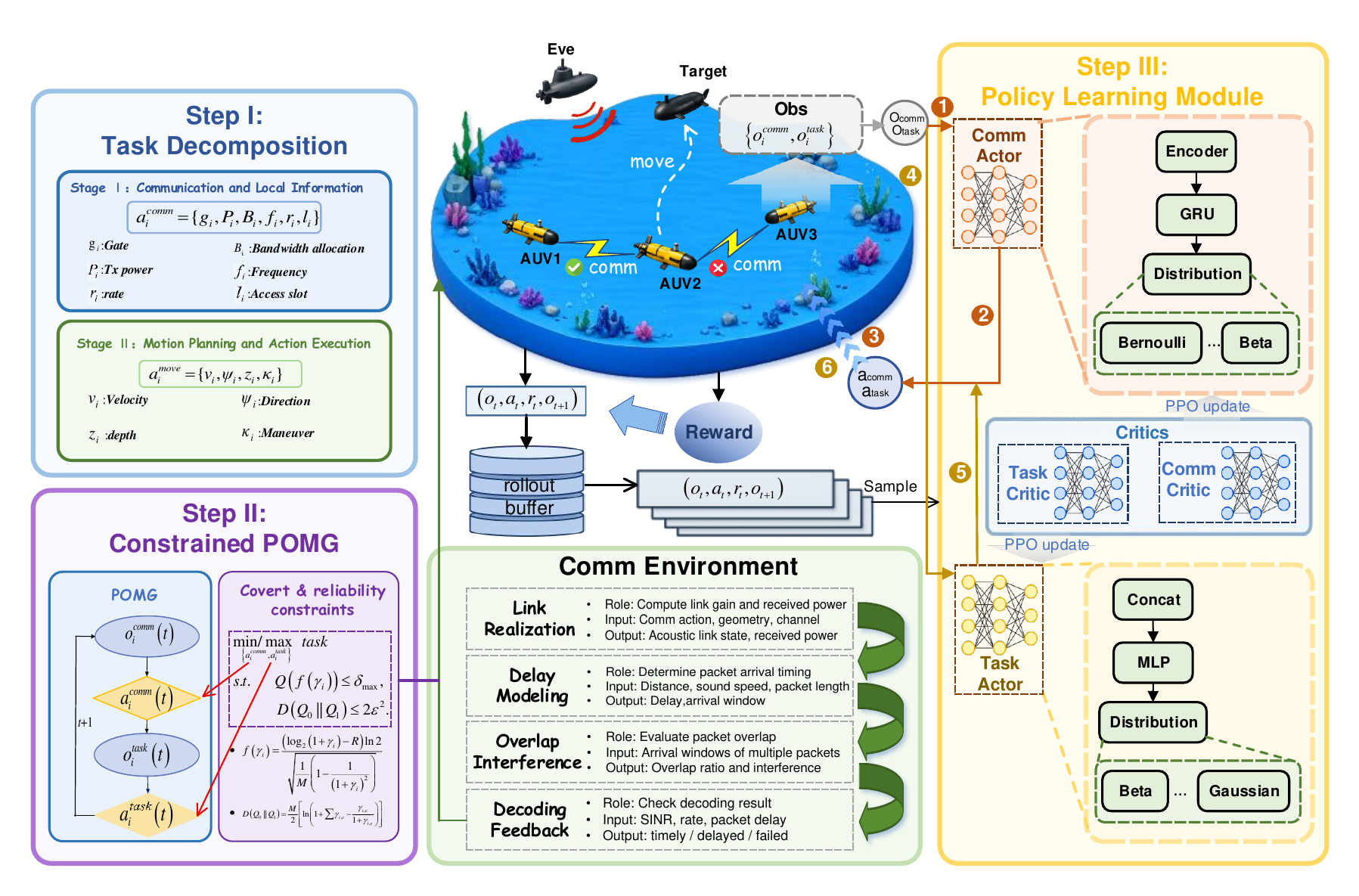}
	\caption{A general framework for covert multi-AUV collaboration.}
    \label{p3}
\end{figure}

\subsection{Value Realization Mechanism of Sensed Information}
\subsubsection{Task-Aware Sensed-Information Message}
This part mainly defines what should be included in the messages transmitted by the AUVs. We divide the message content into three categories.
The first category is \textbf{task observation information}. For example, in a pursuit task, this may include the target bearing, target belief, or local estimation results; in a navigation task, it may include local maps, route risks, or obstacle information.
The second category is \textbf{task intention information}. When AUVs need to perform multi-target pursuit, multi-region search, or cooperative encirclement, the message can include information such as “where I intend to go,” “which target I am responsible for,” or “what role I am currently playing.”
The third category is \textbf{self-state and communication-context information}, such as the AUV’s own position, historical actions, message timestamp, link state, and covert risk. This information can help break the isolated-observation dilemma under fully distributed execution.

On this basis, each AUV combines its own observations, received messages, and recent history to construct a local communication context for transmission decision-making.

\subsubsection{Physics-Grounded Covert Communication Environment}
Unlike many existing MARL frameworks built on Gym-based or simplified simulation environments, where the scenarios often lack realistic communication processes, our proposed framework treats the communication environment as the key realization layer that connects ``value of sensed information'' and ``practical information value''.
Specifically, each AUV first selects a communication action according to its policy. The environment can then account for physical factors such as inter-node distance, channel gain, and underwater acoustic propagation loss to characterize signal propagation and reception. For example, path attenuation can be modeled using the Thorp model \cite{Thorp}, while channel gain can further account for large- and small-scale fading and environmental noise. Based on this, the environment can calculate the receiver-side SINR, message arrival time, transmission duration, and decoding outcome to determine whether teammates can reliably receive the message. In addition, to reflect the practical characteristics of underwater acoustic communications, the environment can further incorporate propagation delay, concurrent interference, and covertness. Because underwater sound travels slowly, packets from different AUVs may arrive asynchronously at the receiver due to differences in propagation distance and transmission duration. When their arrival intervals overlap, concurrent interference can degrade reception quality and decoding performance.

\subsubsection{Practical Information Value-Guided Communication Phase}
As discussed in Section III-A, we have categorized the communication actions. In the communication phase, each AUV generates the corresponding communication action from its local communication context using the communication policy network. For different types of communication decision variables, the communication policy network selects an appropriate action distribution based on whether they are discrete or continuous, and learns state-dependent distribution parameters to generate communication actions. The communication action is then executed in the constructed communication environment, where the extent to which the value of sensed information can be realized is jointly affected by scenario-specific physical channel conditions, communication resource constraints, information timeliness requirements, and covertness constraints.

On this basis, the communication reward $r_{\mathrm{comm}}$ characterizes the practical information value of a message. The reward jointly accounts for the message's task-related benefit and the real communication and covertness costs. Specifically, the task-related benefit represents a message's contribution to team collaboration after it is received and used, and can be measured through changes in the receiver's information state, task-related loss, or team performance. In this context, communication and covertness costs reflect the resources consumed and the exposure risk associated with message transmission. To satisfy system requirements concerning communication reliability, resource budgets, and low detectability, the corresponding conditions can be incorporated into the learning objective as constraints or penalty terms and handled using fixed or adaptive penalties, Lagrangian primal-dual methods, or other constrained reinforcement learning approaches. By jointly evaluating the task-related benefit, communication costs, and associated constraints, the communication policy can balance collaborative benefit against communication feasibility and characterize the net utility that a message can realize under realistic communication conditions.

The task benefit and constraint feedback described above often reflect team-level outcomes and do not directly distinguish individual AUV or message contributions. Contribution-aware credit assignment can therefore associate team benefits and communication costs with the corresponding communication actions. Counterfactual or marginal-contribution methods can estimate the independent effect of an action on collaborative benefit or constraint violation by comparing outcomes with and without that communication action. This reduces ambiguity from uniform collective feedback and provides more targeted learning signals for distributed communication policies. In this way, task-oriented learning signals instantiate the potential collaborative contribution of sensed information as the net utility realizable under practical constraints, while policy optimization closes the loop from the value of sensed information to practical information value.

\subsubsection{Task Phase}
Based on the communication phase results, each AUV updates its local observation and uses the task network to execute physical actions. The specific action form depends on the task and can be represented by velocity, heading, and other variables. The resulting actions further alter the spatial geometry among AUVs, observation quality, channel conditions, and exposure risk, thereby affecting the value realized from communication actions in the next time step. Therefore, in SVR-MARL, the communication and task phases are not independent  but form a closed-loop collaboration process in which “communication improves task decisions, and tasks reshape the communication environment.”

\section{Case Study: Covert Collaborative Localization and Tracking}
\subsection{Scenario and Framework Instantiation}
We consider a three-dimensional underwater collaborative pursuit scenario with a mission area of $1000\mathrm{m} \times 1000\mathrm{m} \times 200\mathrm{m}$, including four cooperative AUVs, one moving target, and one passive eavesdropper, Eve. The AUVs must cooperatively approach and capture the target within a limited time while avoiding frequent or high-power communication. Because of covert requirements, each AUV uses passive sonar as its primary sensing modality. Each AUV can obtain only noisy azimuth and elevation angles of the target relative to itself, without directly measuring range. Moreover, angular observation error increases with target distance, leading to significant uncertainty in a single AUV's target-state estimate.

We build a passive sonar observation model in which each AUV maintains a local belief about the target position, including the estimated position and its uncertainty. At each time slot, an AUV first obtains local angular observations from passive sonar and updates its local belief through an EKF \cite{EKF}. Since a single bearing-only observation is unobservable in range, observations from AUVs at different spatial positions can provide complementary geometric information. Therefore, during the communication phase, an AUV can transmit its sensing message to other AUVs. Once the receiver successfully decodes the message, it updates its target localization estimate with the received information, reducing localization ambiguity from single-agent passive observation.

Each time slot follows the order of “communication phase–movement pursuit phase.” During the communication phase, the communication Actor jointly generates a binary transmission gate $g_i(t)$ and continuous transmit power $p_i(t)$. The former is modeled using a Bernoulli distribution, whereas the latter is modeled using a Beta distribution when the transmission gate is activated. The resulting communication action is then executed in the physical communication environment, where the actual transmission outcome determines reliable message reception and exposure risk. The receiver uses successfully decoded sensing messages to update its target belief. During the movement phase, the task Actor generates continuous movement actions from the updated local belief using a Gaussian policy. Both the communication and task policies are optimized using PPO: the communication policy jointly considers the task-related benefit of a message to cooperative pursuit and the associated communication and covertness costs, whereas the task policy is learned from the pursuit reward. During execution, each AUV makes distributed decisions using only locally available information.

\subsection{Comparative Schemes and Performance Analysis}
\begin{figure}[t]
	\centering
	\includegraphics[width=0.8\columnwidth]{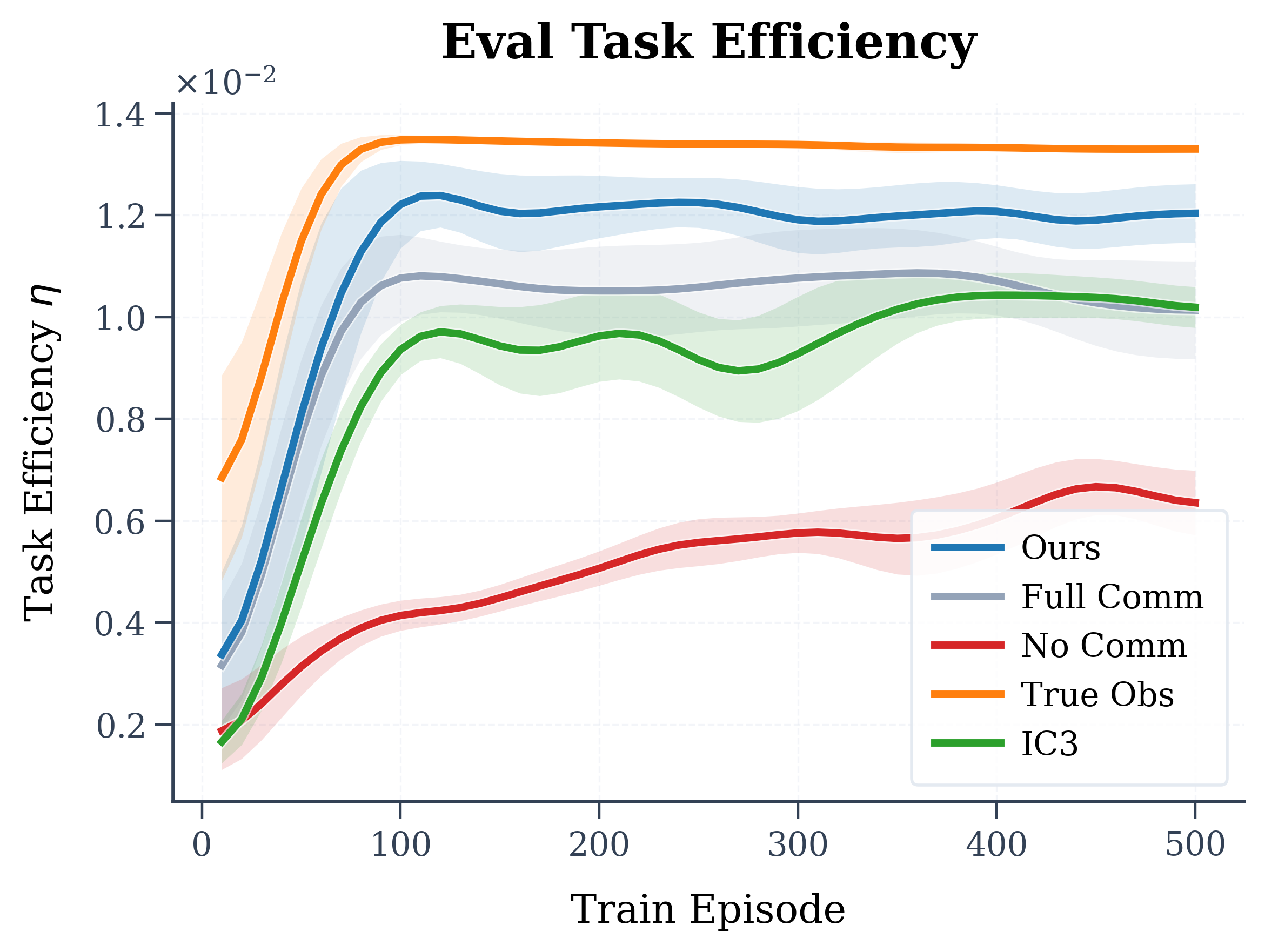}
	\caption{Capture rate and task steps under different communication strategies.}
    \label{exp}
\end{figure}

\begin{figure}[t]
	\centering
	\includegraphics[width=1.02\columnwidth]{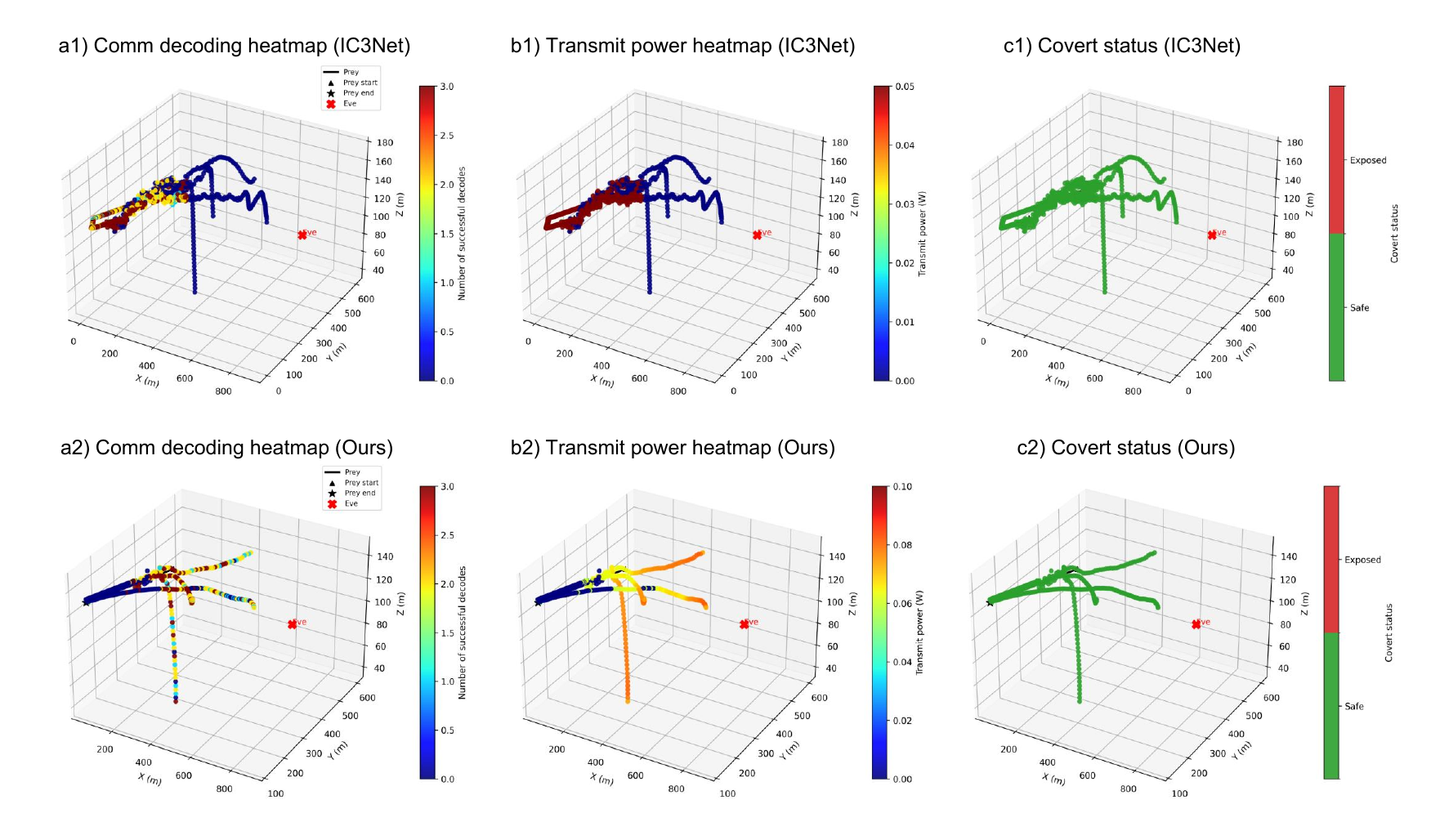}
	\caption{Trajectory-based communication behaviors, including communication decoding, transmit power, and covert status, for IC3Net (a1, b1, c1) and the proposed framework (a2, b2, c2).}
    \label{exp2}
\end{figure}

We consider four comparative schemes:
\begin{itemize}
    \item \textbf{True Obs scheme}: As an upper-bound reference, AUVs obtain dynamic real-time observations of the target without communication, which we use to evaluate the performance of the movement pursuit policy under ideal sensing conditions.
    \item \textbf{Full comm}: Each AUV broadcasts its message at every time slot and transmits with a fixed transmit power that satisfies the covert requirement.
    \item \textbf{No comm}: AUVs do not communicate with each other. Each AUV relies solely on its local observations to update its target belief and complete the movement pursuit task.
    \item \textbf{IC3Net} \cite{IC3}: Each AUV learns a binary communication gate from local observations end-to-end. When the gate activates, it broadcasts the encoded message to teammates through the underwater acoustic channel at fixed transmit power.  
\end{itemize}

Each episode lasts up to $300$ time slots, each slot sequentially executes the communication and movement pursuit phases. The training process employs a dual-policy PPO architecture with alternating updates between the communication and task policies, and the evaluation metric is task efficiency, defined as the ratio of capture rate to task steps.

Fig.~\ref{exp} illustrates the task efficiency achieved by different schemes during evaluation. As the ideal upper bound, True Obs achieves the highest task efficiency by providing accurate target information. In contrast, No Comm performs the worst, indicating that bearing-only passive observations are insufficient for reliable cooperative tracking without information sharing. Full Comm improves task efficiency through AUV communication, but indiscriminate transmissions may introduce channel contention and decoding failures under realistic underwater acoustic links. IC3Net learns communication gating through end-to-end policy optimization, but decoding failures under realistic acoustic and covert constraints can weaken information exchange and cause efficiency fluctuations. Notably, the proposed framework achieves a task efficiency of approximately $1.2\times10^{-2}$, outperforming IC3Net by about 20\% and achieving the best performance among all practical communication schemes.

Fig.~\ref{exp2} further illustrates the communication behaviors of different schemes. Compared with IC3Net, the proposed framework achieves more successful message decodings at critical stages of the task, indicating more effective information sharing among teammates. Meanwhile, under the same covert constraint, it adaptively allocates transmit power according to task requirements, increasing communication investment when target uncertainty is high and reducing unnecessary transmissions once the target is well localized. These results indicate that the proposed framework uses communication resources more efficiently, improving cooperative tracking performance.

\section{Conclusion and Future Direction}
This article proposes a collaborative framework for covert multi-AUV tasks that transforms sensed information into practical information value. Unlike existing multi-agent communication methods that treat communication as an ideal information exchange, this article incorporates realistic underwater acoustic links, propagation delays, concurrent interference, decoding reliability, and Eve-side covert risk into the collaboration loop. In this way, agents can jointly decide whether to communicate and how to communicate, based on both task value and physical realizability. In the case study, we consider a covert multi-AUV collaborative pursuit task and verify that value-driven communication scheduling can effectively improve target localization and task completion efficiency under bearing-only passive observations. The results show that the proposed framework improves multi-AUV collaboration performance under realistic communication and covert constraints, offering a new design perspective for task-oriented optimization of underwater covert communication.


\subsubsection{Teammate Intention Prediction for Distributed Coordination}
Although we adopt a fully distributed execution architecture, each AUV makes communication and task decisions based on local observations and received messages. However, without global scheduling, an AUV may still struggle to accurately infer its teammates' future communication behaviors, movement directions, and task roles. This may lead to repeated transmissions, overly conservative decisions, or action conflicts. In future work, we will consider introducing a teammate-intention prediction module that enables each AUV to recursively estimate other AUVs' communication tendencies and action intentions over the next few steps, based on historical messages and communication records. In this way, AUVs can better use communication messages to predict the team's collaborative state based on local information and thus decide how to communicate more effectively. This direction is expected to further improve scheduling efficiency and task coordination in distributed covert collaboration.

\subsubsection{Secure Agent Authentication and Trustworthy Communication}
In real underwater environments, even if the communication policy controls acoustic exposure risk, the system may still face security threats, such as malicious node attacks or false-message injection. An attacker may impersonate a legitimate AUV and send incorrect messages to disrupt team collaboration. Future work can further investigate secure authentication and trustworthy message evaluation mechanisms based on acoustic fingerprints, channel characteristics, motion behaviors, and historical communication patterns. These mechanisms can distinguish legitimate nodes from malicious ones and assess message credibility before fusion, thereby improving the robustness and communication security of underwater distributed cooperative systems in complex adversarial environments.

\vspace{11pt}

\vfill

\end{document}